%% file: main.tex
\pdfoutput=1

\documentclass[11pt]{article}

\usepackage[preprint]{acl}

\usepackage{arabtex}
\usepackage{utf8}
\setcode{utf8}

\usepackage{times}
\usepackage{latexsym}
\usepackage[T1]{fontenc}
\usepackage[utf8]{inputenc}
\usepackage{microtype}
\usepackage{textcomp}
\usepackage{upquote}
\usepackage{inconsolata}
\usepackage{wrapfig}
\usepackage{dblfloatfix}
\usepackage{placeins}
\usepackage{titlesec}
\usepackage{float}
\usepackage{enumitem}

\usepackage{amsmath,amssymb,amsfonts}
\usepackage{algorithmic}
\usepackage{mathscinet}
\usepackage{graphicx}

\usepackage{xcolor}
\usepackage{color, colortbl}

\definecolor{LightGrey}{rgb}{0.8,0.8,0.8}
\definecolor{LightGrey2}{rgb}{0.95,0.95,0.95}

\definecolor{heatGreen}{HTML}{1E4620}
\definecolor{heatLightG}{HTML}{E2F0D9}
\definecolor{heatLightR}{HTML}{FCE4D6}
\definecolor{heatRed}{HTML}{C00000}

\usepackage{tcolorbox}
\usepackage{caption}
\usepackage{subcaption}
\usepackage{array}
\usepackage{multirow}
\usepackage{booktabs}
\usepackage{tabularray}
\usepackage{ninecolors}
\UseTblrLibrary{booktabs}

\usepackage{url}
\usepackage{hyperref}

\renewenvironment{abstract}%
  {\begin{center}\large\textbf{\abstractname}\end{center}%
    \begingroup\leftskip=0.6cm\rightskip=0.6cm\noindent\ignorespaces}%
  {\par\endgroup}

\titlespacing*{\section}      {0pt}{1.6ex plus .3ex minus .2ex}{0.6ex plus .15ex}
\titlespacing*{\subsection}   {0pt}{1.1ex plus .25ex minus .15ex}{0.4ex plus .1ex}
\titlespacing*{\subsubsection}{0pt}{0.9ex plus .2ex minus .1ex}{0.3ex plus .1ex}
\titlespacing*{\paragraph}    {0pt}{0.6ex plus .15ex}{0.6em}

\makeatletter
\AddToHook{begindocument/before}{%
  \gdef\@fileswithoptions#1{%
    \@ifnextchar[{\@fileswith@ptions#1}{\@fileswith@ptions#1[]}}}
\makeatother

\title{ArEssays: Towards Robust Arabic Handwritten Text Understanding\\ in Educational Settings}

\author{
  Hamdy Mubarak, Anas Madkoor, Ehsaneddin Asgari, Majd Hawasly \\
  Qatar Computing Research Institute, Hamad Bin Khalifa University, Doha, Qatar \\
  \texttt{\{hmubarak,amadkoor,easgari,mhawasly\}@hbku.edu.qa}}

\begin{document}
\maketitle

\begin{abstract}
Automated understanding of handwritten Arabic student essays remains challenging due to script complexity and high variability in children’s writing. We introduce \textbf{ArEssays}, a large-scale dataset of authentic essays from primary to secondary students across Egypt, with both verbatim transcriptions and linguistically corrected versions. We benchmark 34 vision-language and OCR models, including fine-tuned systems, and find large performance gaps on real student handwriting. We release the dataset, training pipelines, and results to support research in Arabic OCR, document understanding, and educational NLP.

\end{abstract}

\section{Introduction}
\input{sections/introduction}

\section{Related Work}
 \input{sections/related_work}

\section{Corpus Construction}
\input{sections/corpus_construction}

\section{Corpus Analysis}
\input{sections/corpus_analysis}

\section{Models}
\input{sections/models}

\section{Results}
\label{sec:results}
\input{sections/results}

\section{Experiments}
\label{sec:Experiments}
Using the same evaluation setup and prompts (Appendix~\ref{sec:gen_benchmark_prompt}) across the ArEssays dataset partitions comprising 500 training, 100 development, and 200 testing instances.

\section{Conclusion}
We introduced \textbf{ArEssays}, the first multi-paragraph essay-level corpus of authentic Arabic handwriting produced by school children, with paired verbatim and linguistically corrected ground truth. Our benchmark of thirty-four state-of-the-art Vision-Language Models, Arabic-specialized systems, and fine-tuned baselines establishes a strong, reproducible foundation for Arabic educational document intelligence and exposes a clear gap between current models and the demands of authentic developmental handwriting. We further introduced a two-stage LoRA recipe that explicitly decouples visual specialization from linguistic regularization, yielding the strongest results on ArEssays at three distinct parameter scales.

Beyond the immediate benchmark, our findings highlight a methodological lesson for the wider community: when ground-truth annotation contains the very kind of noise that pretrained models implicitly correct, a single error metric is insufficient. The dual-annotation framework and $\Delta$CER analysis introduced here provide a concrete blueprint that we expect to be useful in any educational, dialectal, or historical-text setting where the distinction between \emph{what was written} and \emph{what was meant} carries real-world consequence. We release the corpus, training pipelines, evaluation harness, and full benchmark results to support follow-up work on Arabic OCR, educational NLP, and multimodal language understanding. An extended discussion of the implications of our findings for Arabic NLP, educational AI, and multimodal learning is provided in \S\ref{app:discussion}.

\section*{Limitations}
Although the proposed corpus provides a large-scale and realistic benchmark for Arabic handwritten student essays, several limitations should be acknowledged. First, the dataset was collected exclusively from students in Egypt and therefore primarily reflects Egyptian educational practices, handwriting conventions, linguistic usage, and sociocultural writing patterns. While the corpus includes students from five governorates and covers both male and female students across grades 4–12, the findings may not generalize to other Arabic-speaking countries, dialects, educational systems, demographic groups, or socioeconomic settings. Moreover, the corpus has grade imbalance (Figure \ref{fig:essay_per_grade}) and gender imbalance (Figure \ref{fig:gender_by_stage}). Consequently, broader conclusions regarding Arabic handwriting, OCR robustness, or educational language development should be validated through additional studies conducted under different geographical and educational conditions.

Second, although the essays were collected under naturally occurring school and home conditions using mobile phone cameras, the corpus may still not capture the full range of real-world document variability. Extremely degraded images, severe blur, truncated pages, and highly illegible handwriting were partially filtered during corpus construction to ensure annotation quality and usability. Future releases may include more challenging samples to support robustness evaluation under harsher visual conditions.

Third, the corrected versions of the essays were manually produced by expert linguists, and while annotators were instructed to preserve students’ intended meaning and writing style, certain normalization decisions may introduce subjective bias, particularly in cases involving ambiguous spelling, punctuation, or sentence segmentation. Similarly, sentence-level segmentation in Arabic handwritten text can be inherently ambiguous due to inconsistent punctuation and spacing behavior in student writing.

Fourth, we acknowledge potential sources of error in the corpus and analysis pipeline. Automatic linguistic analysis tools, including CAMeL Tools and Stanza, may introduce morphological and syntactic analysis errors, particularly for noisy handwritten student text. Furthermore, manual transcription, correction, and metadata annotation may contain occasional human annotation errors, especially in subjective tasks such as image-quality assessment and handwriting readability evaluation.

Fifth, the OCR and vision-language benchmark results reported in this study are dependent on the selected evaluation models, prompts, preprocessing decisions, and decoding settings. As OCR and multimodal language models continue to evolve rapidly, benchmark performance may change substantially with future systems.

Finally, although the corpus enables analysis of linguistic development across educational stages, the dataset is cross-sectional rather than longitudinal. The essays therefore represent different students at different grade levels rather than tracking the same students over time. As a result, the observed developmental trends should be interpreted cautiously.

\section*{Ethical Considerations}
\label{sec:ethical_considerations}
\begin{figure}
    \centering
\includegraphics[width=0.7\linewidth]{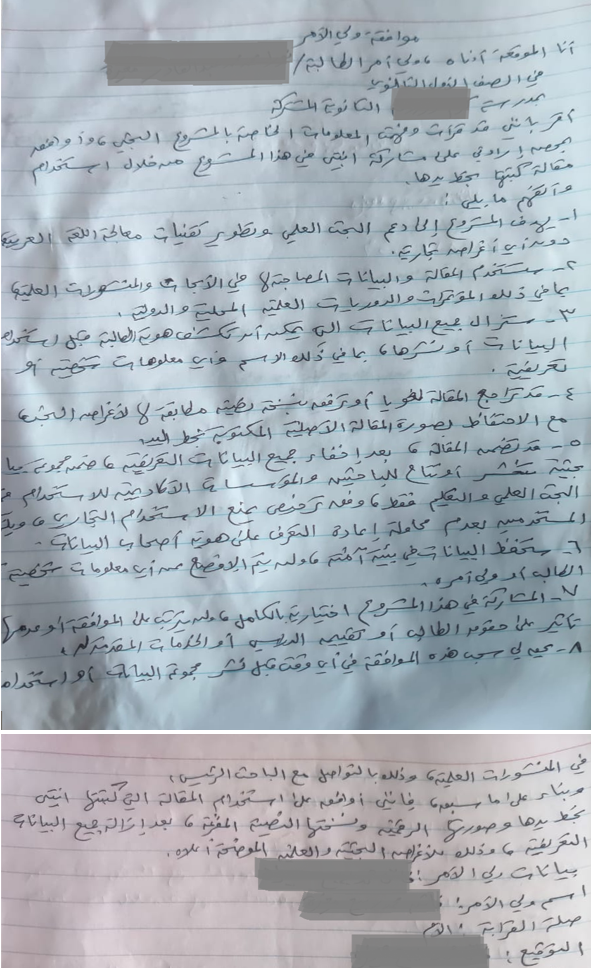}
    \caption{Parental Consent Form. Parents or legal guardians provided informed consent for students to participate in collecting handwritten Arabic essays for research on Arabic NLP, handwriting recognition, and educational technologies. Participation was voluntary and had no impact on students’ academic assessment. Collected data included essay images and transcriptions, grade level, and optional demographic information. Essays were manually reviewed to remove personally identifiable information, and participants were assigned anonymous IDs. Original data were securely stored with access restricted to the research team. An anonymized version of the dataset may be publicly released for research purposes. Parents could withdraw consent before the dataset’s publication.\\
}
    \label{fig:parent_consent}
\end{figure}

This work involves handwritten essays produced by school students and therefore raises important ethical and privacy considerations. Prior to participation, parental consent was obtained for all participating students (sample in Figure \ref{fig:parent_consent}). The data collection process was coordinated through school teachers, and participation was voluntary.

To protect participant privacy, all released data were anonymized. The corpus does not contain personally identifying information about students, and anonymized identifiers were used for both students and teachers. Metadata associated with the released corpus is limited to non-identifying attributes such as educational grade, gender, and governorate.

The essays were collected under natural educational conditions in which students wrote by hand without the use of digital devices, spell checkers, or AI-based writing assistants. Teachers captured the essays using mobile phone cameras to simulate realistic classroom and home acquisition settings commonly encountered in Arabic OCR applications.

The handwritten essays were manually transcribed and corrected by expert linguists. The correction process focused on fixing spelling, orthographic, and punctuation errors while preserving the students’ intended meaning and writing style as much as possible. Because educational writing may contain developmental language patterns and non-standard forms, annotators were instructed to avoid excessive rewriting or stylistic normalization.

Participants were compensated fairly for their time and effort. Students received approximately USD 10 per essay, while teachers were compensated at a rate above typical local annotation and educational task rates for coordinating collection and image acquisition activities. We intentionally adopted compensation practices designed to avoid exploitative data collection conditions.

The released corpus, benchmark code, and evaluation results will be shared for research purposes to support reproducibility and future work in Arabic OCR, educational NLP, and multimodal language understanding. However, we acknowledge that handwriting datasets involving minors may still carry potential privacy risks despite anonymization. Accordingly, the dataset is intended strictly for academic and research use, and users are expected to follow responsible data usage practices.

\textbf{Using AI Tools:} We used ChatGPT solely for language refinement and improving the clarity of the manuscript writing. It was not used for generating scientific content, creating datasets, producing annotations, or generating experimental results. In addition, Claude Code was used to assist in implementing Python scripts for data analysis and visualization. All analysis functions and processing steps were independently verified through unit testing before generating the reported tables and figures.

\bibliography{references}

\clearpage
\appendix
\section{Samples}

\begin{figure}
    \centering
\includegraphics[width=1.0\linewidth]{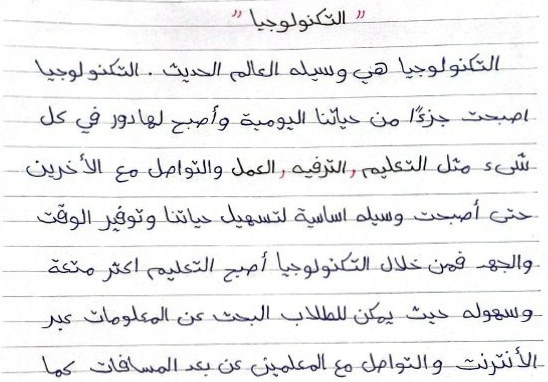}
    \caption{Sample easy essay. Metadata: Government: Giza, Grade: 6, Gender: Female, Handwritten clarity: 3 (Very clean handwritten), Visual noise: 3 (Clean), Image quality: 3 (High), Topic: Technology.\\
}
    \label{fig:essay_sample_org2}
\end{figure}

\begin{figure}
    \centering
\includegraphics[width=1.0\linewidth]{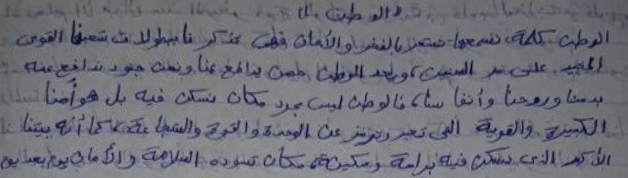}
    \caption{Sample hard essay. Metadata: Government: Damietta, Grade: 12, Gender: Male, Handwritten clarity: 1 (Hard to read), Visual noise: 1 (Heavy noise), Image quality: 1 (Low/blurry), Topic: Homeland.\\
}
    \label{fig:essay_sample_org3}
\end{figure}

Figures \ref{fig:essay_sample_org2} and \ref{fig:essay_sample_org3} show samples of the essays with highest and lowest measures for handwritten clarity, noise, and image quality.

\section{Top Essay Topics}
\label{sec:essay_topics}
\begin{figure}
    \centering
\includegraphics[width=1.0\linewidth]{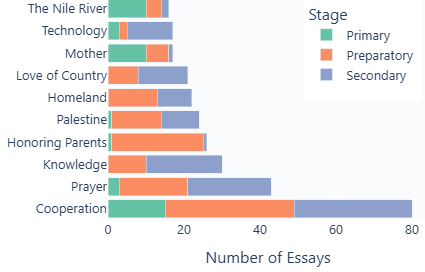}
    \caption{Top 10 essay topics by stage}
    \label{fig:essay_topics}
\end{figure}

Figure \ref{fig:essay_topics} shows the top 10 essay topics by stage.

\section{Corpus Collection Details}
\label{sec:corpus_collection_details}
participating students are from grades 4 (G4) through 12 (G12) as students from lower primary grades (e.g., Grades 1–3) typically produce isolated words or short phrases rather than complete handwritten essays.

 Participants in the corpus building were compensated for their time and effort. Students received approximately USD 10 per essay, while teachers were compensated at a rate of USD 8 per hour for coordination, collection, and scanning activities. In addition, Expert linguists were compensated at a rate of USD 8 per hour of linguistic annotations. Compensation rates were intentionally set above typical local annotation and educational task rates to ensure fair participation and sustained engagement throughout the project. All teachers and linguists are native Arabs from Egypt with experience ranging between 20-30 years.

In total, the collection process produced 1,200 handwritten essays. In the current release, we publicly provide 800 essays along with their associated annotations and metadata. The remaining essays were excluded from the current release after manual quality assessment due to factors such as: low image resolution, poor lighting conditions, excessive visual artifacts, or highly illegible handwriting.

Rather than discarding these difficult samples permanently, we plan to include many of them (with annotations and metadata) in future releases as part of a challenging benchmark for robust Arabic OCR and vision-language systems operating under realistic degradation conditions.

The collection process took around three months while the annotation processes (transcription, correction, metadata enrichment, quality control) took seven months.

\subsection{Post-processing}
Handwritten student essays were manually post-processed using the \textbf{CamScanner} application to improve image usability while preserving the original handwriting content. The processing pipeline included automatic perspective correction, rotation and angle adjustment, background cropping, and removal of surrounding non-document objects such as desks and classroom artifacts. This step was necessary because essays were captured using mobile phone cameras under natural classroom conditions, resulting in substantial variation in orientation, framing, and background clutter.

\section{Error Rate}
\label{sec:error_rate}
\begin{figure}
    \centering
\includegraphics[width=1.0\linewidth]{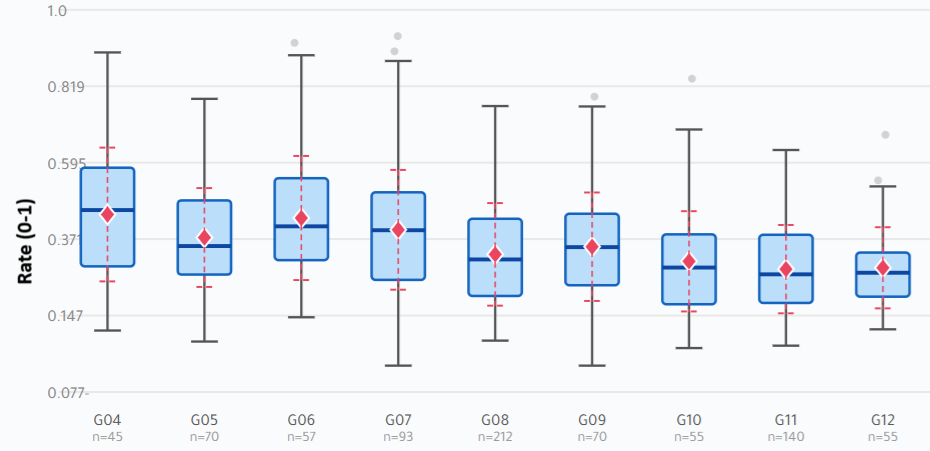}
    \caption{Error rate by grade}
    \label{fig:error_rate}
\end{figure}

Figure \ref{fig:error_rate} shows error rate by grade.

\section{Error Mapping}
\label{sec:error_mapping}
\begin{figure}
    \centering
\includegraphics[width=1.0\linewidth]{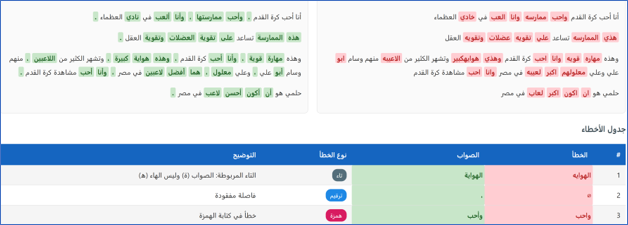}
    \caption{Example of error mapping and classification}
    \label{fig:error_mapping}
\end{figure}

Figure \ref{fig:error_mapping} shows an example after aligning original words and corrected words. Right is the transcribed essay image, and left is the corrected text. Errors are colored and classified into Taa/Haa, punctuation, and Hamza errors.

\section{Writing style Analysis}
\label{sec:writing_style}
The essays reveal naturally occurring educational and sociolinguistic writing patterns. Lower-grade students frequently rely on memorized narrative templates and repetitive introductory expressions, while secondary-school students demonstrate greater rhetorical variation, thematic organization, and discourse coherence. Formulaic expressions such as religious quotations, standardized openings, and memorized concluding phrases occur frequently across educational stages, reflecting authentic Egyptian school-writing conventions. The progression from simple narrative writing in primary school to more descriptive and argumentative writing in secondary school highlights the corpus’s value for studying Arabic writing development under realistic educational conditions.

\section{Experimental Setup}
\label{sec:experimental_setup}

\textbf{Experimental Setup.} All training and experiments were conducted on a machine with the specifications detailed in Table~\ref{tab:exp_setup}. A total of approximately 98 GPU hours were utilized across all six training runs (three models $\times$ two stages). Stage~1 runs were brief (0.3--1.8 hours each), while Stage~2 runs were substantially longer, averaging approximately 32 hours per model, with individual runs ranging from 19 to 50 hours.

\begin{table}[!h]
\small
    \centering
    \begin{tabular}{c|c}
    \hline
      \textbf{CPU}   & INTEL(R) XEON(R) PLATINUM 8568Y \\
      \rowcolor{LightGrey2}
      \textbf{GPU}   &  Nvidia H200-141GB\\
      \textbf{Memory} & 512GB\\
      \rowcolor{LightGrey2}
      \textbf{OS} & Ubuntu 20.04.5 LTS\\
      \hline
     \end{tabular}
    \caption{Specification of machine used for training and experiments}
    \label{tab:exp_setup}
\end{table}

\section{Benchmark Evaluation Prompt and Decoding Settings}
\label{sec:gen_benchmark_prompt}

For all benchmarked off-the-shelf systems described in \S\ref{ssec:benchmarked}, we adopt a single fixed prompt and a unified set of decoding parameters to ensure a fair, reproducible zero-shot comparison across model families. The exact prompt template and decoding configuration used at evaluation time are reproduced in this appendix.

\begin{tcolorbox}[colback=blue!5!white, colframe=blue!75!black,
  title={Zero-Shot Transcription Prompt --- Qwen3.5-4B (Test159 / Article0468)}]

\textbf{System Message:} \\
You are an OCR engine. Output ONLY the text visible in the image.
Do not think, reason, or add explanations. \\

\textbf{User Message (Exact OCR task):} \\
Transcribe the handwritten Arabic text in this image exactly as it appears,
character by character, word by word.
Do NOT correct any spelling or grammar mistakes.
Output ONLY the transcribed text. \\

\textbf{Input Image:} \\
\includegraphics[width=6.5cm]{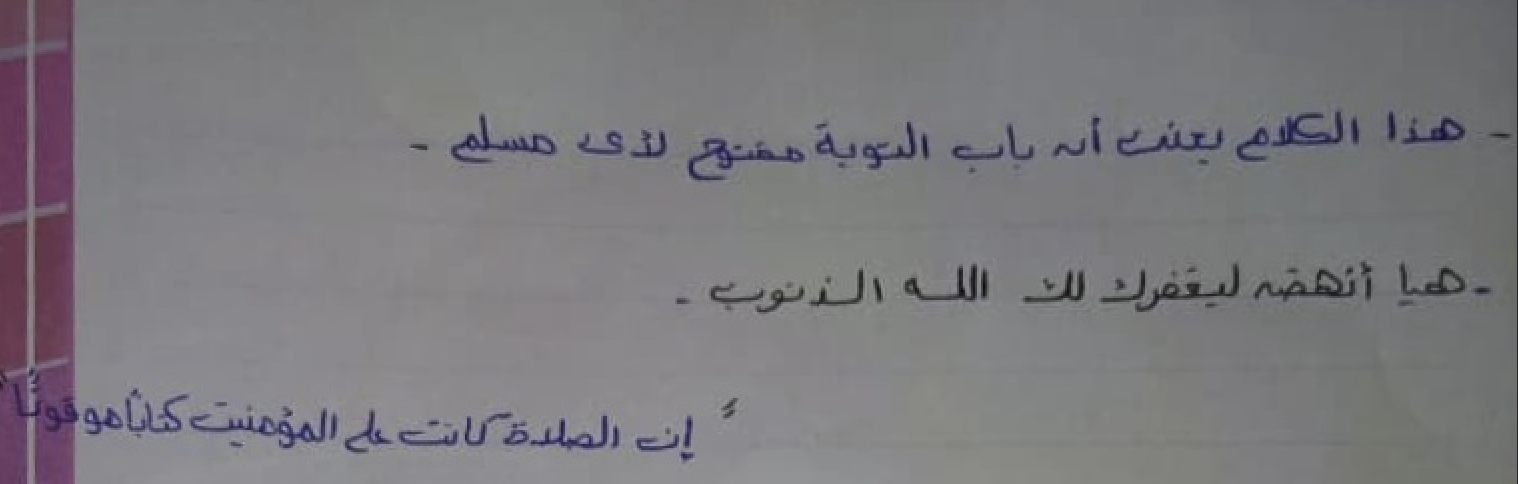} \\

\textbf{Ground Truth:} \\
\tiny
\begin{RLtext}- هذا الكلام يعنى أن باب التوبة مفتوح لأي مسلم. - - هيا أنهض ليغفرك لك الله الذنوب -. «إن الصلاة كانت على المؤمنين كتابا موقوتا».\end{RLtext}

\textbf{Model Output (Qwen3.5-4B):} \\
\tiny
\begin{RLtext}- هذا الكلام يعنى أن باب التوبة مفتوح لأي مسلم. - - هيا انهض ليغفر لك الله الذنوب -. «إن الصلاة كانت على المؤمنين كتابا موقوتا».\end{RLtext}

\textbf{Scores:} \\
\textbf{Exact CER:} 0.0625 \hfill \textbf{Exact WER:} 0.1818

\end{tcolorbox}

\noindent The corresponding decoding parameters applied uniformly to every benchmarked checkpoint are listed in Table~\ref{tab:decoding_params}. Where a model exposes additional task-specific switches (e.g., dedicated OCR or layout modes), the exact values used are reported in Table~\ref{tab:decoding_model_specific}.

\begin{table}[t]
    \centering
    \small
    \setlength{\tabcolsep}{0pt}
    \begin{tabular}{l|l}
    \toprule
    \textbf{Decoding parameter} & \textbf{Value} \\
    \midrule
    Decoding strategy       & Greedy search (\texttt{do\_sample=False}) \\
    \rowcolor{LightGrey2}
    Max new tokens          & 1{,}024 \\
    Repetition penalty      & model-specific$^{\dagger}$ \\
    \rowcolor{LightGrey2}
    Beam size               & 1 \\
    Random seed             & N/A (deterministic) \\
    \rowcolor{LightGrey2}
    Image preprocessing     & Resize to max resolution, RGB convert \\
    Max image resolution    & model-specific$^{\dagger}$ \\
    \bottomrule
    \end{tabular}
    \caption{Common decoding parameters used across all benchmarked models
    (\S\ref{ssec:benchmarked}).
    $^\dagger$Per-model overrides are listed in
    Table~\ref{tab:decoding_model_specific}.}
    \label{tab:decoding_params}
\end{table}
\begin{table*}[t]
    \centering
    \small
    \setlength{\tabcolsep}{8pt} %
    \begin{tabular}{l p{12.0cm}}
    \toprule
    \textbf{Model} & \textbf{Model-specific setting} \\
    \midrule
    Qwen2.5-VL / Qwen3-VL
        & No system prompt; standard greedy generate \\
    \rowcolor{LightGrey2}
    Qwen3.5 / Qwen3.6
        & System prompt: \textit{``You are an OCR engine\ldots''}; \texttt{enable\_thinking=False} \\
    Gemma-3 family
        & No additional flags; default \texttt{generate()} \\
    \rowcolor{LightGrey2}
    olmOCR
        & \texttt{rep.\ penalty=1.15}; \texttt{no\_repeat\_ngram=20} \\
    PaddleOCR-VL
        & Prompt override: ``\texttt{OCR:}''; \texttt{rep.\ penalty=1.15}; \texttt{no\_repeat\_ngram=20} \\
    \rowcolor{LightGrey2}
    dots.ocr
        & Layout-JSON prompt; custom KV-cached greedy decode; text extracted from JSON \\
    GLM-OCR
        & Prompt override: ``\texttt{Text Recognition:}''; \texttt{rep.\ penalty=1.15}; \texttt{no\_repeat\_ngram=20} \\
    \rowcolor{LightGrey2}
    Baseer
        & \texttt{max\_pixels=$644{\times}644$}; \texttt{rep.\ penalty=1.1} \\
    Qari-OCR
        & \texttt{rep.\ penalty=1.1} \\
    \rowcolor{LightGrey2}
    AIN
        & \texttt{max\_pixels=$1{,}280{\times}1{,}280$}; standard greedy \\
    Fanar (Oryx)
        & \texttt{max\_pixels=$1{,}280{\times}1{,}280$}; \texttt{rep.\ penalty=1.05} \\
    \rowcolor{LightGrey2}
    Ketaba-OCR-LoRA
        & \texttt{max\_pixels=$1{,}280{\times}28{\times}28$}; \texttt{rep.\ penalty=1.1} \\
    \bottomrule
    \end{tabular}
    \caption{Model-specific configuration applied at evaluation time.
    All other decoding parameters follow Table~\ref{tab:decoding_params}.
    \textit{rep.\ penalty} = repetition penalty;
    \textit{no\_repeat\_ngram} = \texttt{no\_repeat\_ngram\_size}.}
    \label{tab:decoding_model_specific}
\end{table*}

\section{Fine-Tuning Hyperparameters}
\label{sec:training_hyperparams}

This appendix details the full set of optimization and adaptation
hyperparameters used in the two-stage \mbox{LoRA} recipe described in
\S\ref{ssec:finetuned}.
Table~\ref{tab:hparams_common} lists settings shared across all runs.
Table~\ref{tab:hparams_stage1} reports Stage~1 (Target-Domain
Specialization, \textbf{v1}) and Table~\ref{tab:hparams_stage2} reports
Stage~2 (Curricular Regularization, \textbf{v2}), both broken down per
base architecture.

\begin{table}[H]
    \centering
    \small
    \begin{tabular}{l|l}
    \toprule
    \textbf{Setting} & \textbf{Value} \\
    \midrule
    Optimizer            & AdamW (fused) \\
    \rowcolor{LightGrey2}
    Precision            & bfloat16 \\
    LR scheduler         & cosine with warm-up \\
    \rowcolor{LightGrey2}
    Warm-up ratio        & 0.05 \\
    Weight decay         & 0.05 \\
    \rowcolor{LightGrey2}
    Gradient clipping    & 1.0 (max norm) \\
    LoRA target modules  & all linear projection layers \\
    \rowcolor{LightGrey2}
    LoRA dropout         & 0.05 \\
    Max sequence length  & 1{,}024 tokens \\
    \rowcolor{LightGrey2}
    Random seed          & 42 \\
    \bottomrule
    \end{tabular}
    \caption{Common training settings shared across all fine-tuning runs
    (Stage~1 and Stage~2, all architectures).
    Max image resolution varies per architecture and is listed in
    Tables~\ref{tab:hparams_stage1}--\ref{tab:hparams_stage2}.}
    \label{tab:hparams_common}
\end{table}

\begin{table*}[t]
    \centering
    \small
    \begin{tabular}{l|c|c|c}
    \toprule
    \textbf{Hyperparameter}
        & \textbf{Baseer-3B}
        & \textbf{Qwen3.5-4B}
        & \textbf{Gemma4-26B-A4B} \\
    \midrule
    LoRA rank ($r$)              & 64                       & 64                         & 64 \\
    \rowcolor{LightGrey2}
    LoRA $\alpha$                & 128                      & 128                        & 128 \\
    Learning rate                & $3{\times}10^{-5}$       & $8{\times}10^{-5}$         & $2{\times}10^{-5}$ \\
    \rowcolor{LightGrey2}
    Effective batch size         & 16                       & 16                         & 16 \\
    Per-device batch size        & 2                        & 1                          & 1 \\
    \rowcolor{LightGrey2}
    Gradient accumulation steps  & 8                        & 16                         & 16 \\
    Number of epochs (max)       & 15                       & 12                         & 15 \\
    \rowcolor{LightGrey2}
    Early-stop patience          & 4 epochs                 & 3 epochs                   & 4 epochs \\
    Training steps (actual)      & 466                      & 351                        & 465 \\
    \rowcolor{LightGrey2}
    Max image resolution         & $644{\times}644$         & $2{,}048{\times}28^2$\,px  & $1{,}024{\times}28^2$\,px \\
    Augmentation probability     & 0.50                     & 0.50                       & 0.50 \\
    \rowcolor{LightGrey2}
    Trainable parameters         & 160.5\,M                 & 78.1\,M                    & 50.6\,M \\
    GPU hours ($1{\times}$H100)  & 0.72\,h                  & 1.80\,h                    & 0.34\,h \\
    \bottomrule
    \end{tabular}
    \caption{Stage~1 (\textbf{v1}: Target-Domain Specialization)
    fine-tuning hyperparameters on the ArEssays training partition.
    Early stopping halted Baseer at epoch~7 and Qwen3.5 at epoch~6;
    Gemma4 completed all 15 epochs.
    See \S\ref{ssec:finetuned}.}
    \label{tab:hparams_stage1}
\end{table*}

\begin{table*}[t]
    \centering
    \small
    \begin{tabular}{l|c|c|c}
    \toprule
    \textbf{Hyperparameter}
        & \textbf{Baseer-3B}
        & \textbf{Qwen3.5-4B}
        & \textbf{Gemma4-26B-A4B} \\
    \midrule
    Stage-1 adapter merging      & yes                      & yes                        & yes \\
    \rowcolor{LightGrey2}
    LoRA rank ($r$)              & 32                       & 32                         & 16 \\
    LoRA $\alpha$                & 64                       & 64                         & 32 \\
    \rowcolor{LightGrey2}
    Learning rate                & $1{\times}10^{-5}$       & $5{\times}10^{-5}$         & $5{\times}10^{-6}$ \\
    Effective batch size         & 16                       & 16                         & 16 \\
    \rowcolor{LightGrey2}
    Per-device batch size        & 2                        & 2                          & 2 \\
    Gradient accumulation steps  & 8                        & 8                          & 8 \\
    \rowcolor{LightGrey2}
    Number of epochs (max)       & 5                        & 10                         & 8 \\
    Training steps (actual)      & 14{,}811                 & 6{,}697                    & 23{,}611 \\
    \rowcolor{LightGrey2}
    Training corpus & \multicolumn{3}{c}{KHATT \citep{Mahmoud2014KHATT} $+$ Books-OCR \citep{seemorg_books_ocr}} \\
    Max image resolution         & $644{\times}644$         & $1{,}024{\times}28^2$\,px  & $1{,}024{\times}28^2$\,px \\
    \rowcolor{LightGrey2}
    Augmentation probability     & 0.20                     & 0.20                       & 0.20 \\
    Trainable parameters         & 79.5\,M                  & 156.1\,M                   & ${\approx}$12.7\,M \\
    \rowcolor{LightGrey2}
    GPU hours ($1{\times}$H100)  & 18.8\,h                  & 25.8\,h                    & 50.5\,h \\
    \bottomrule
    \end{tabular}
    \caption{Stage~2 (\textbf{v2}: Curricular Regularization)
    fine-tuning hyperparameters.
    Stage-1 adapters are merged into the base weights before training;
    a fresh lower-rank adapter is then trained on KHATT
    \citep{Mahmoud2014KHATT} combined with a printed Books-OCR
    collection \citep{seemorg_books_ocr}.
    All three models completed their full epoch budget.
    See \S\ref{ssec:finetuned}.}
    \label{tab:hparams_stage2}
\end{table*}

\section{Extended Discussion and Implications}
\label{app:discussion}

Our benchmark surfaces three findings whose implications extend beyond Arabic Handwritten Text Recognition. This appendix expands on each of them, referenced from the Conclusion in the main body.

\noindent\textbf{Implications for Arabic NLP.}
On authentic children's handwriting, broad multimodal scaling alone is not sufficient: the largest non-adapted Vision-Language Models still trail dedicated fine-tuned 3B-scale models by a substantial margin. Targeted, in-domain data dominates parameter count in this regime, echoing observations in low-resource Arabic NLP more broadly \cite{Nacar2025SARD, Hennara2025Baseer}. The negative $\Delta$CER exhibited by large general-purpose VLMs further reveals that they conflate visual transcription with implicit linguistic correction, silently smoothing developmental orthographic errors towards Modern Standard Arabic. While this behaviour is valuable for downstream language understanding, it is a hazard for any system whose ground truth is meant to be the \emph{actually written} script, the precise setting our dual-annotation protocol is designed to disentangle.

\noindent\textbf{Implications for educational AI.}
Automated educational technology in Arabic-speaking contexts cannot be built on transcripts that silently correct what students wrote, since these are exactly the signals teachers and learning-analytics systems need to track developmental progress. Our results show that a separation between \emph{verbatim} and \emph{linguistically corrected} targets is essential for building trustworthy formative-assessment pipelines: verbatim transcription should drive error analysis and feedback, while corrected transcription supports downstream NLU. The clarity-$\times$-grade stratification in \S\ref{sec:results} further indicates that handwriting clarity, not grade level alone, is the principal bottleneck, so investments in image capture, paper standardization, and lightweight pre-processing in classrooms are likely to yield larger gains than further raw model scaling.

\noindent\textbf{Implications for multimodal learning.}
The two-stage LoRA curriculum, Stage~1 visual specialization followed by Stage~2 curricular regularization on KHATT and printed text, yields the strongest models on ArEssays and recovers performance across grade and clarity strata. The explicit decoupling of \emph{visual specialization} from \emph{linguistic regularization} provides a transferable recipe for any multimodal task that combines a narrow visual distribution with a broad linguistic prior. We hypothesize the same recipe will be effective in adjacent settings such as historical Arabic manuscripts, low-resource cursive scripts (Persian, Urdu), and child-produced texts in other languages, and we release our training code, prompts, and adapters to facilitate this line of work.

\end{document}

%% file: sections/introduction.tex
Robust Handwritten Text Recognition (HTR) is essential for educational AI systems for grading, learning analytics, and language analysis. While Vision–Language Models (VLMs) perform well on printed and adult written English text, Arabic handwriting especially children’s evolving script remains challenging \cite{Nacar2025SARD, Hennara2025Baseer, Ayadi2025CNN}. This limits scalable educational technology in Arabic speaking contexts.

Arabic student handwriting combines intrinsic script complexity with developmental noise. Arabic is cursive and position-dependent through ligatures and dotting (\emph{I'jam}); children further introduce unstable baselines, inconsistent sizing, erratic diacritics, and irregular spacing. Classroom data also includes mobile-phone capture artifacts, varied writing tools, different paper types, and erasures. Consequently, OCR systems trained on synthetic, printed, or adult handwriting \cite{Heakl2025KITAB, Mahmoud2014KHATT} often fail on authentic student prose through attention drift, character substitutions, and linguistic hallucinations.

Existing children's Arabic HTR efforts \cite{Alheraki2022HandwrittenACB, Durayhim2023TowardsACD, Alwagdani2023DeepLCK, Alshehri2025ImageBasedROF} are restricted to \emph{isolated character} recognition under largely controlled acquisition conditions, with a single transcription target per sample. They therefore cannot characterize how modern VLMs behave on multi-paragraph essays, where ligature distortions, baseline drift, orthographic variability, and developmental spelling errors interact. ArEssays introduced in this paper fills this gap.

The shift to multi-paragraph, multi-grade essays exposes a methodological subtlety: when ground truth contains orthographic noise that pretrained models tend to smooth away, standard Word/Character Error Rate becomes ambiguous. A model may lower CER by correcting the student rather than transcribing faithfully. For educational applications, this distinction should be explicit, which we address through paired \emph{verbatim} and \emph{linguistically corrected} ground truth.\\

\noindent\textbf{Contributions.} Our contributions in this work are:
\begin{enumerate}[leftmargin=*,itemsep=2pt,topsep=2pt]
    \item[(I)] One of the largest known Arabic children's writing collections, with grade-stratified essays from Grades~4 through~12 written under naturally occurring classroom conditions.
    \item[(II)] A \emph{dual ground-truth} annotation protocol providing both \textit{verbatim} and \textit{linguistically corrected} targets, enabling a clean separation of visual transcription from grammatical correction.

    \item[(III)] A detailed analysis of morphological and syntactic development, as well as error patterns, across students at different educational stages.

    \item[(IV)] A comprehensive benchmark of \textbf{thirty-four} models: general-purpose VLMs, document-specialized OCR engines, Arabic-specialized systems, and our own fine-tuned baselines.
    \item[(V)] A two-stage Low-Rank Adaptation (LoRA) training recipe that decouples visual specialization from linguistic regularization, with public adapters at three parameter scales.
    \item[(VI)] A detailed error analysis correlating Character/Word Error Rates with educational stage, handwriting clarity, and visual-noise strata, surfacing actionable findings for Arabic educational AI.
\end{enumerate}
\noindent
Code, data, and trained recipes will be released for academic use. %

%% file: sections/related_work.tex
\noindent\textbf{Arabic Handwritten Character and Text Recognition.}
Arabic HTR has evolved from Gabor/SIFT/HMM pipelines \cite{ElHajj2005ArabicHRAC, Mohamad2009CombiningSCAD, Elzobi2012ArabicHRBA, Rabi2018RecognitionOCBC} to CNN classifiers for isolated characters \cite{Younis2017ArabicHCAK, Altwaijry2020ArabicHRA, Wagaa2022ImprovedAAL, Kamal2022HurufAAM, Attar2025DeepCNH}, CNN, BiLSTM/ConvLSTM, and CTC sequence models \cite{Eltay2020ExploringDLS, Gader2022AttentionbasedCFN, Dahbali2024AHMAP}, and Transformer recognizers \cite{Kassem2021OCFormerATAO, Kassem2022AnEOU, Momeni2023ATAR, Chan2024HATFormer}. End-to-end Arabic document OCR pipelines \cite{Mosbah2024ADOCRNetADAR, Waly2024ArabicHDP, Waly2025InvizoAHV} and broad surveys \cite{Lorigo2006OfflineAHBB, Balaha2020AutomaticROAM, Alrobah2022ArabicHRAA, Khayati2024CNNbasedMFAH, Kasem2023AdvancementsACBK} document this trajectory. All of these systems are trained predominantly on printed, historical, or adult sources, leaving authentic classroom data largely uncovered.

\noindent\textbf{Arabic Handwriting Corpora and the Children Gap.}
KHATT \cite{Mahmoud2014KHATT} remains the canonical paragraph-level adult benchmark; Muharaf \cite{Saeed2024MuharafMOAG} and HICMA \cite{Ismail2023HICMATHAN} target cursive and calligraphic manuscripts; ADBase/MADBase \cite{ElSherif2007ATSO} support digit recognition. The few studies addressing children's Arabic handwriting \cite{Alheraki2022HandwrittenACB, Durayhim2023TowardsACD, Alwagdani2023DeepLCK, Alshehri2025ImageBasedROF, Qedear2024RealTimeARE} are confined to \emph{isolated characters} under controlled conditions, with a single transcription target. CER-HV \cite{AlAzzawi2026CERHVAHAI} further motivates human-in-the-loop curation for Arabic HTR. To our knowledge, no prior public benchmark targets multi-paragraph essay-level handwriting from Arabic school children across grade bands with paired verbatim and corrected ground truth, the gap that ArEssays fills.

\noindent\textbf{Vision-Language Models for Arabic Document Intelligence.}
End-to-end Arabic document understanding is now dominated by multimodal LLMs and VLMs. Qalam \cite{Bhatia2024QalamAMAF} introduced an Arabic-tailored multimodal LLM; KITAB \cite{Heakl2025KITAB}, Baseer \cite{Hennara2025Baseer}, and SARD \cite{Nacar2025SARD} provide benchmarks and Arabic-adapted backbones; community efforts such as Qari-OCR \cite{Wasfy2025QariOCR} and AIN \cite{Heakl2025AIN} extend the Qwen-VL family with synthetic Arabic OCR data. While these systems achieve strong results on printed and adult-produced material, their robustness on authentic in-the-wild educational handwriting is largely uncharacterized, the evaluation gap that ArEssays directly addresses.

\noindent\textbf{Educational NLP and Learner-Language Analysis.}
Beyond pure recognition, our work relates to Arabic educational NLP, including the ZAEBUC writer corpus \cite{habash-palfreyman-2022-zaebuc}, Arabic Automated Essay Scoring systems using synthetic data and CEFR-aligned error modeling \cite{qwaider-etal-2025-aes}, and Arabic Grammatical Error Correction methods based on noisy learner text \cite{alhafni-habash-2025-gec}. However, these approaches assume clean typed transcriptions, which is unrealistic for handwritten input. Our dual-annotation framework explicitly separates verbatim transcription from linguistic correction, bridging OCR and downstream educational NLP and enabling end-to-end handwritten Arabic AES, GEC, and learner-language analysis.

%% file: sections/corpus_construction.tex
\label{sec:corpus_contruction}
\subsection{Data Collection}

To construct the corpus, we collaborated with two school teachers in Egypt who coordinated the data collection process. These teachers subsequently recruited eight additional teachers from five Egyptian governorates from differnt geographical regions to ensure diversity in handwriting styles, educational backgrounds, and regional writing variations.

Prior to participation, parental consent was obtained for all participating students (Details in \ref{sec:ethical_considerations}). The collection protocol was designed to preserve natural handwriting conditions while maintaining consistency across contributors.

Students from grades 4 (G4) through 12 (G12) were asked to handwrite an essay on one of 15 general-domain topics. The topics were intentionally selected to be broad, age-appropriate, and familiar to students from different educational levels in order to encourage free-form writing and natural language production. Most students contributed a single essay, typically consisting of approximately one handwritten page.
To ensure authentic writing conditions, students wrote their essays in the presence of their teachers without using digital devices, spelling correction systems, or AI-based writing assistants. After completion, teachers captured the essays using mobile phone cameras under naturally occurring classroom and home conditions, resulting in realistic variation in lighting, resolution, orientation, and image quality.

 In the current release, we publicly provide 800 essays along with their associated annotations and metadata. Details of the corpus collection and post processing steps can be found in the Appendix \ref{sec:corpus_collection_details}.

\subsection{Annotation and Text Processing}

Each essay was manually processed by four expert Arabic linguists through multiple annotation stages.
\begin{itemize}[leftmargin=*, noitemsep, topsep=0pt]
    \item \textbf{Line-Level Transcription}: The handwritten essays were first transcribed line-by-line while preserving the original line boundaries appearing in the image. This alignment enables research on handwritten OCR, line segmentation, and multimodal text-image alignment.

\item \textbf{Text Correction}: A corrected version of each essay was subsequently produced by expert linguists. The correction process focused on fixing spelling, orthographic, and punctuation errors while preserving the student’s intended meaning and writing style as much as possible.

\item \textbf{Sentence-Level Segmentation}: In addition to the line-level transcription, linguists created a sentence-segmented version of each essay in which text was reorganized into complete semantic sentences rather than image-based line boundaries. This representation supports downstream NLP tasks such as language modeling, educational assessment, OCR post-correction, and syntactic analysis.
\end{itemize}

\subsection{Metadata}
Each essay is accompanied by a set of metadata fields, including: student grade level (Grades 4–12), gender, essay title, governorate, and anonymized student and teacher information to protect privacy.

\subsection{Image Quality Assessment}
Because handwriting legibility and image quality substantially affect OCR performance, all essays were independently evaluated by two linguists using three quality dimensions. The average inter-annotator agreement across the quality dimensions was approximately 80\%. Disagreements were resolved through consolidation discussions. The quality measures are defined as follows:\\
\noindent
\textbf{I. Legibility}: This criterion evaluates the readability of the handwriting itself and serves as the most important quality factor. \textbf{Ratings:} 3-Very clear handwriting, 2-Legible with moderate effort, and 1-Difficult to read.\\
\noindent
\textbf{II. Visual Noise}: This criterion captures the presence of artifacts such as corrections, strike-throughs, decorations, smudges, background clutter, and additional markings. \textbf{Ratings:} 3-Clean image, 2-Some visual noise, and 1-Heavy visual noise.\\
\noindent
\textbf{III. Image Quality}:
This criterion evaluates technical image characteristics including sharpness, brightness, contrast, and resolution. \textbf{Ratings:} 3-High-quality image, 2-Medium quality, and 1-Low or blurry image.

\subsection{Corpus Statistics}
\begin{table}[H]
\centering
\small
\begin{tabular}{lllr}
\toprule
\textbf{Target} & \textbf{Attribute} & \textbf{Values} & \textbf{\%} \\
\midrule
Student & Stage  &  Primary (G4-G6)     & 22   \\
       & &  Preparatory (G7-G9) & 47 \\
       & & Secondary (G10-G12) & 31 \\
\rowcolor{LightGrey2}
Student & Gender & Female & 57 \\
\rowcolor{LightGrey2}
       & & Male & 32 \\
\rowcolor{LightGrey2}
       & & Unspecified & 11\\
Student & Government & Giza (GC, Center) & 22\\
           & & Cairo  (GC, Center) & 12 \\
           & & Qalyubia (GC, Center) & 10\\
           & & Damietta (North East) & 43 \\
           & & Qena (UE, South) & 13\\
\rowcolor{LightGrey2}
\midrule
Image & Handwritten & Very clear & 26\\
\rowcolor{LightGrey2}
       & clarity & Legible with effort & 56 \\
\rowcolor{LightGrey2}
      & & Hard to read & 18\\
Image & Visual noise & Clean & 26 \\
      & & Some noise & 60 \\
      & & Heavy noise & 14 \\
\rowcolor{LightGrey2}
Image & Quality & High & 6\\
\rowcolor{LightGrey2}
      & & Medium & 79\\
\rowcolor{LightGrey2}
      & & Low/blurry & 15\\
\bottomrule
\end{tabular}
\caption{Corpus Statistics. Number of essays = 800.\\
GC: Greater Cairo, UE: Upper Egypt.}
\label{tab:corpus_statistics}
\end{table}

\begin{figure}[t]
    \centering
    \includegraphics[width=0.95\linewidth]{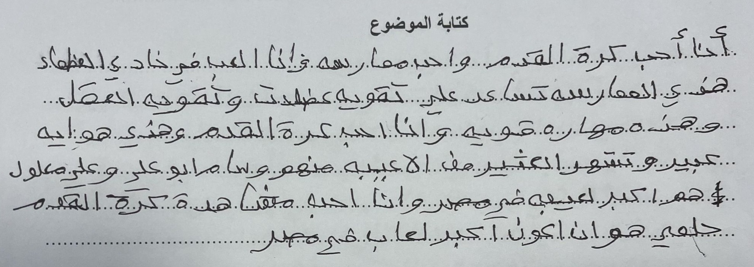}
    \caption{Sample essay. Metadata: Government: Giza, Grade: 6, Gender: Male, Handwritten clarity: 2 (Legible with effort), Visual noise: 3 (Clean), Image quality: 2 (Medium), Topic: Hobby. Translation: ``I love football, I love playing it, and I play for the club of greats. This practice helps strengthen both my muscles and \dots''}
    \label{fig:essay_sample_org}
\end{figure}

\begin{figure}
    \centering
\includegraphics[width=1.0\linewidth]{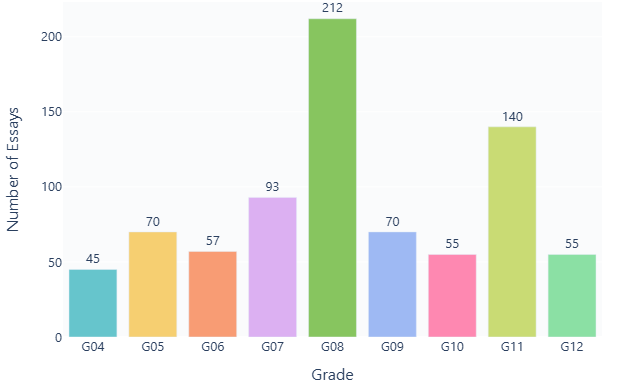}
    \caption{Distribution of essays by grade}
    \label{fig:essay_per_grade}
\end{figure}

\begin{figure}
    \centering
\includegraphics[width=1.0\linewidth]{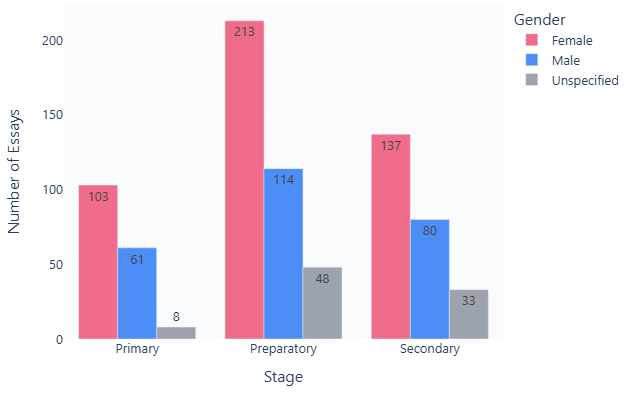}
    \caption{Distribution of gender by stage}
    \label{fig:gender_by_stage}
\end{figure}

Figure \ref{fig:essay_sample_org} shows a sample of the essays with its metadata. Other samples with different quality measures can be found in the Appendix (Figures \ref{fig:essay_sample_org2} and \ref{fig:essay_sample_org3}). Figure \ref{fig:essay_per_grade} shows the number of essays by grade while Figure \ref{fig:gender_by_stage} shows the gender distribution of students by school stage, and Figure \ref{fig:essay_topics} in the appendix shows the top 10 essay topics.

%% file: sections/corpus_analysis.tex
\label{sec:corpus_analysis}
After segmenting the student essays into sentence-level logical units, we performed morphological analysis using CAMeL Tools\footnote{ \url{https://github.com/CAMeL-Lab/camel_tools}} and syntactic analysis using Stanza NLP\footnote{\url{https://stanfordnlp.github.io/stanza/}} on the corrected sentences.

To analyze writing and OCR-related errors, the original transcription and corrected essay versions were aligned at the word level using \textbf{SequenceMatcher} from Python’s \textbf{difflib} package. Each aligned word pair was then compared to identify the underlying error category based on character-level differences. For example, substitutions involving Alif (\<ا>) and Hamza (\<أإآ>) forms were categorized as \textbf{Hamza errors}, while confusions between Taa Marbouta (\<ة>) and Haa (\<ه>) were labeled as \textbf{Taa/Haa errors}, and confusions between Yaa (\<ي>) and Alif Maqsoura (\<ى>) were mapped to \textbf{Yaa errors}. Differences in whitespace usage (e.g., long spacing between letters in a word) were classified as \textbf{spacing errors}, and punctuation mismatches were grouped under \textbf{punctuation errors}. In addition, grammar-related errors were identified through morphologically motivated substitutions, such as confusion between grammatical case and agreement endings (e.g., \<ون> vs \<ين>, or Tanween), which were categorized as \textbf{grammatical errors}. Other errors involving missing letters, added letters, letters written in odd shapes or angles (e.g., writing \<نـ> as \<خـ>)\footnote{Please refer to Figure \ref{fig:essay_sample_org} for examples.}, style errors (e.g., using dialectal and informal words), etc. were mapped to \textbf{Others/Unknown}.

\begin{figure}
    \centering
\includegraphics[width=1.0\linewidth]{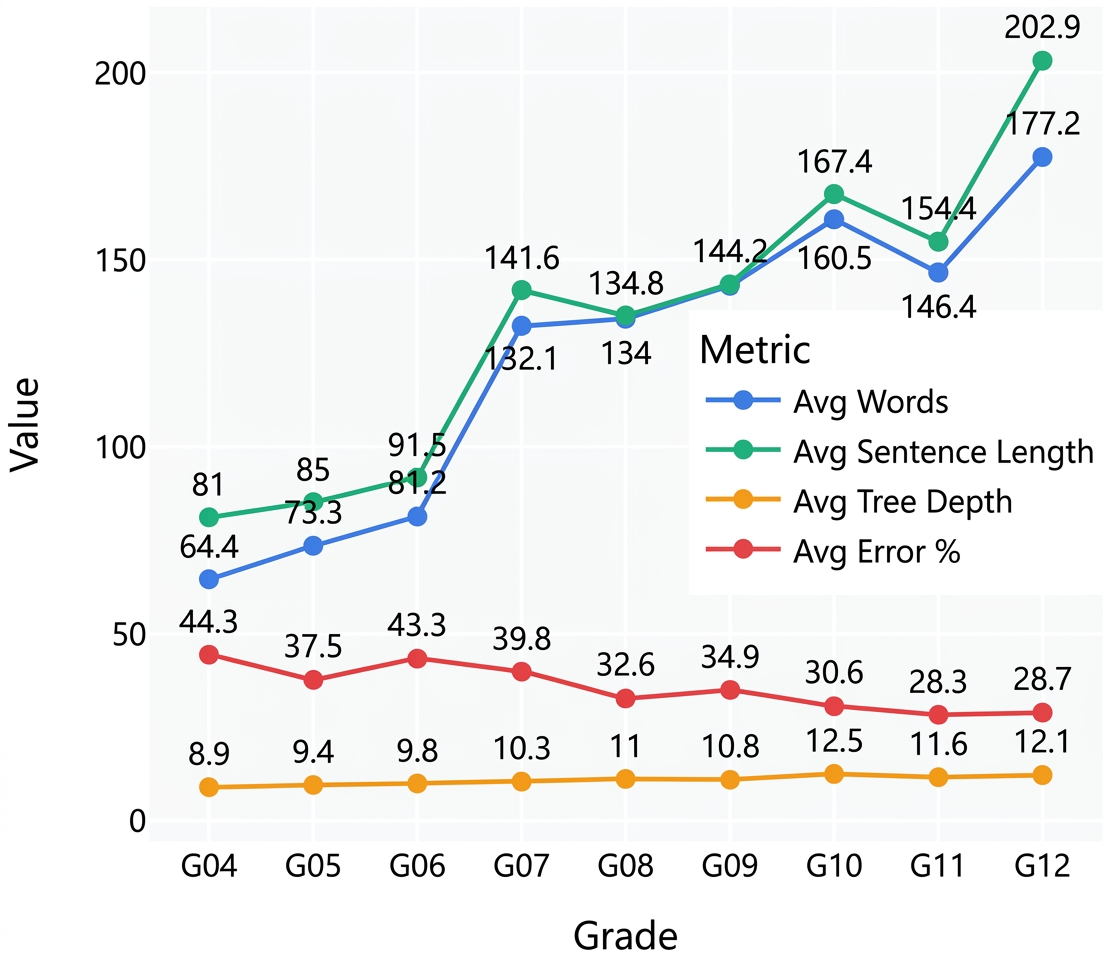}
    \caption{Key metric averages by grade}
    \label{fig:key_metric}
\end{figure}

Figure \ref{fig:key_metric} summarizes the key metric averages by grade while the detailed linguistic statistics and analyses are presented in the following subsections. Figure \ref{fig:error_mapping} in the appendix shows an example of error mapping and classification.

\subsection{Linguistic Analysis}
The corpus exhibits substantial linguistic diversity across grades 4–12, capturing developmental variation in Arabic writing proficiency, lexical richness, morphology, syntax, discourse organization, and orthographic behavior. The handwritten essays were produced under authentic educational conditions, enabling analysis of naturally occurring student writing across primary, preparatory, and secondary educational stages.

\begin{figure}
    \centering
\includegraphics[width=1.0\linewidth]{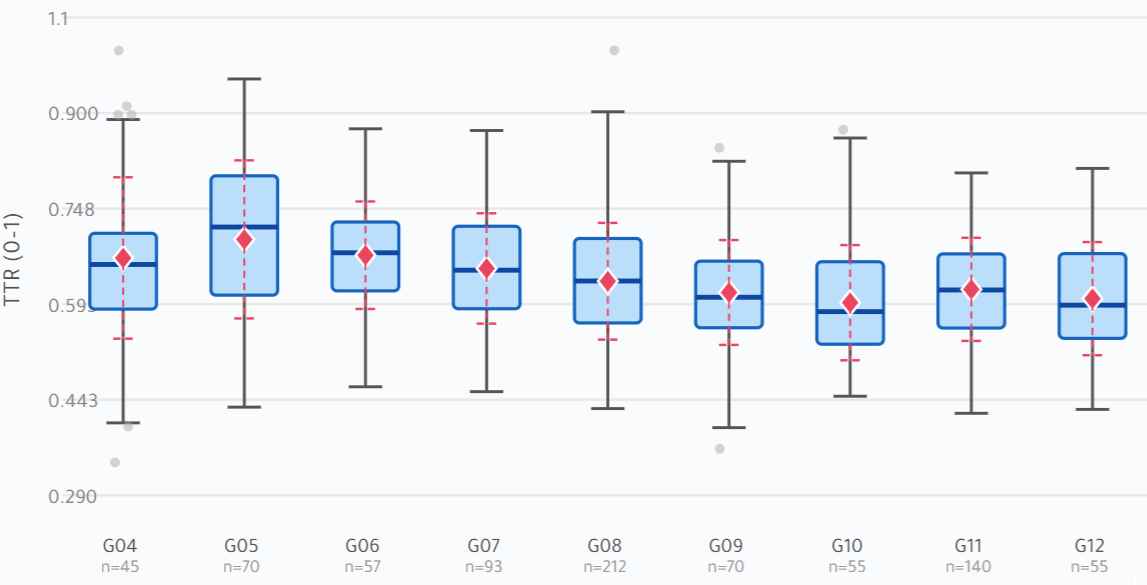}
    \caption{Type-Token-Ratio (TTR) by grade}
    \label{fig:ttr}
\end{figure}

\textbf{Vocabulary Development:} Lexical analysis shows clear progression in vocabulary richness across grades. Primary essays (Grades 4–6, average 69 words/essay) use limited, repetitive vocabularies (40–70 unique lemmas; TTR 0.48–0.56) and rely on simple concrete terms. Preparatory essays (Grades 7–9, average 124 words/essay) show moderate growth (80–120 lemmas) with more descriptive language and discourse connectives. Secondary essays (Grades 10–12, average 149 words/essay) are lexically richer, often exceeding 150 lemmas and including more abstract and domain-specific vocabulary (Figure~\ref{fig:ttr}). Average word length also increases from 3.8–4.2 characters in lower grades to over 5 in secondary stages, reflecting higher morphological and semantic complexity.

\begin{figure}
    \centering
\includegraphics[width=1.0\linewidth]{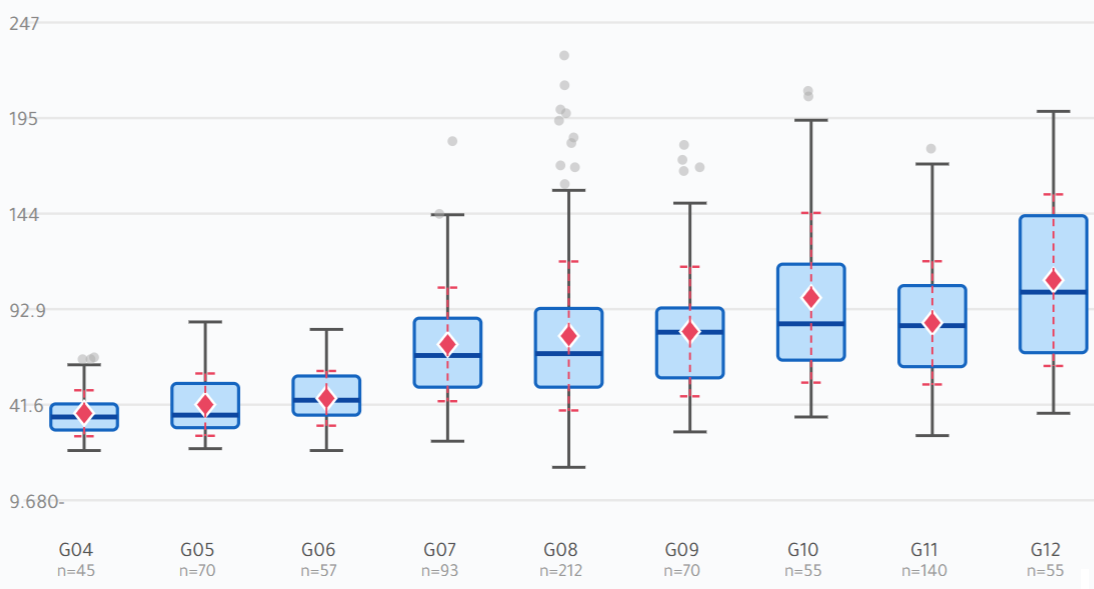}
    \caption{Number of clitics by grade}
    \label{fig:clitics}
\end{figure}

\textbf{Morphology:} Morphologically, the corpus captures the richness and complexity of Arabic writing through extensive inflectional variation, clitic attachment, and derivational morphology. The corpus contains substantial clitic attachment phenomena, with clitic attachment rates exceeding 0.55 in many essays and surpassing 0.80 in some secondary-school samples. These characteristics create substantial challenges for OCR tokenization and downstream Arabic NLP systems. (Figure \ref{fig:clitics})

\begin{figure}
    \centering
\includegraphics[width=1.0\linewidth]{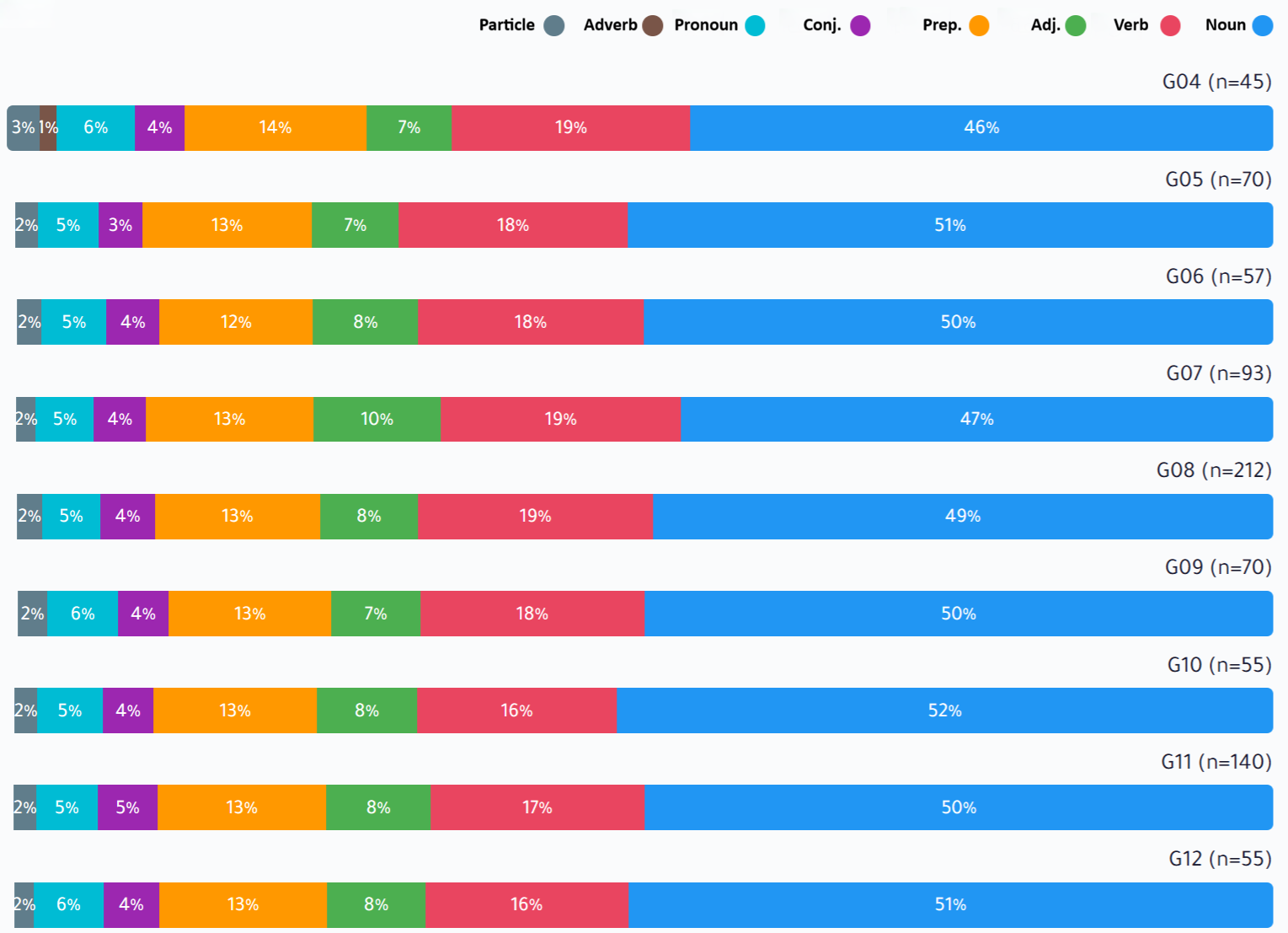}
    \caption{Parts of Speech (POS) by grade}
    \label{fig:pos}
\end{figure}

 Nouns constitute the dominant part-of-speech category across all grades, representing 46–52\% of total tokens, while verbal usage varies depending on essay topic and educational level. Primary-school essays predominantly contain simple verbal constructions and repetitive sentence templates, often relying heavily on present-tense narration. In contrast, secondary-school essays demonstrate more balanced use of past and present tense, richer adjectival modification, and increasingly complex nominal structures (Figure \ref{fig:pos}).

\begin{figure}
    \centering
\includegraphics[width=1.0\linewidth]{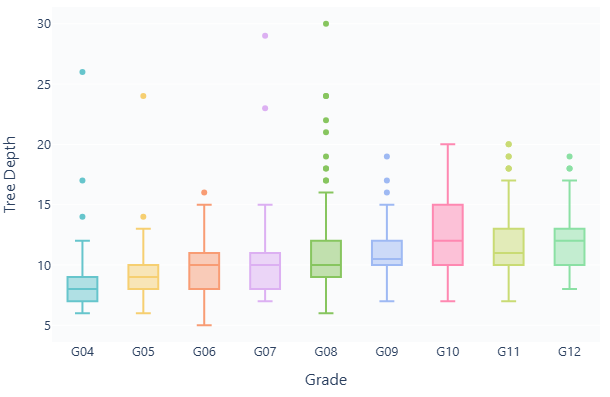}
    \caption{Parse-tree depth by grade}
    \label{fig:parse_trees}
\end{figure}

\textbf{Syntax:} Syntactic complexity similarly increases with grade level. Primary-school essays typically contain short sentences with shallow dependency structures, limited subordination, and relatively weak discourse organization. Average sentence lengths in Grades 4–6 are frequently below 10–12 tokens per sentence, and many essays contain minimal punctuation and sentence segmentation cues. Preparatory-school essays demonstrate noticeable improvement in sentence organization, connective usage, and clause structure. Secondary-school essays, however, frequently contain long multi-clause constructions with deeper dependency trees and more sophisticated argumentative organization. Several Grade 11–12 essays exhibit dependency depths exceeding 12 levels and subordinate clause ratios above 0.20. Some essays additionally contain extremely long sentences exceeding 300 tokens due to narrative chaining and inconsistent punctuation behavior. (Figure \ref{fig:parse_trees})

\textbf{Error Types:} The corpus exhibits extensive orthographic and punctuation variation affecting OCR and downstream NLP. Common issues include Hamza omission, Taa Marbouta/Haa confusion, Alif variants, spacing inconsistencies, and missing punctuation, especially in lower grades. Hamza errors persist across all stages, with some secondary essays requiring 30+ corrections. Punctuation is often sparse or inconsistent, requiring near-complete restoration. The data also includes segmentation issues such as merged words, irregular spacing, and fused character boundaries, which are challenging for Arabic OCR. Figure~\ref{fig:error_types} shows error distributions by grade, while Figure~\ref{fig:error_rate} in the appendix reports error rates from 44\% (Grade 4) to 29\% (Grade 12). Writing style details are provided in Appendix~\ref{sec:writing_style}.

Overall, the corpus integrates linguistic and developmental variation, handwriting diversity, and OCR challenges in a single resource. Its combination of visual noise, orthographic ambiguity, and writing variability makes it useful for Arabic OCR, handwriting recognition, OCR post-correction, educational NLP, multimodal document understanding, and modeling Arabic writing development.

\begin{figure}
    \centering
\includegraphics[width=1.0\linewidth]{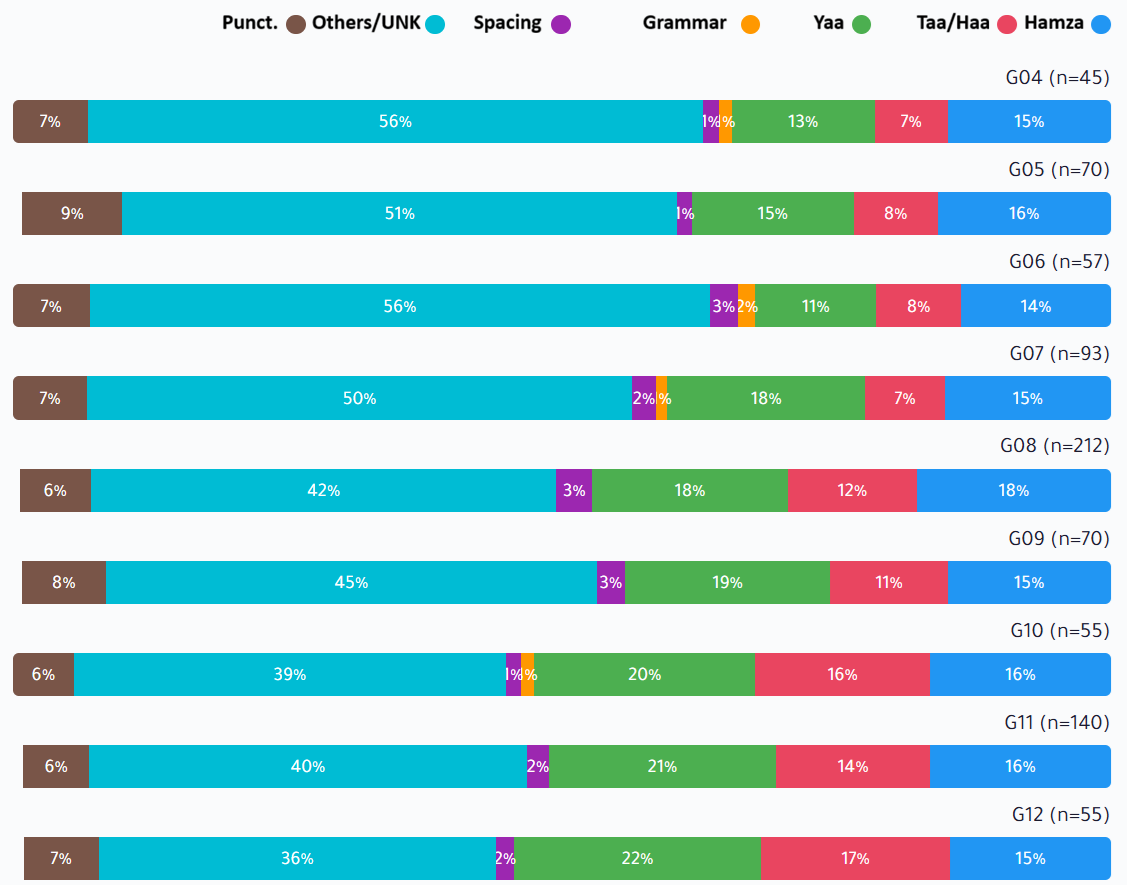}
    \caption{Error types by grade}
    \label{fig:error_types}
\end{figure}

\subsection{Linguistic Correction (Correction OCR)}
The Correction OCR task couples visual text extraction with grammatical error correction (GEC) and orthographic normalization. Given a handwritten image $X$, the model generates a corrected sequence $\hat{Y}_c$ aligned with expert-annotated ground truth $Y_c$, which preserves meaning while fixing grammar and orthography \cite{Heakl2025KITAB}. The task evaluates whether models can resolve visual ambiguities, restore missing diacritics, and normalize dialectal Egyptian Arabic into standard MSA.

\subsection{Layout Segmentation}
Document layout analysis and line level segmentation remain critical bottlenecks for continuous paragraph processing. In this benchmark, the objective is to localize and isolate individual text lines and word bounding structures within full page unconstrained essay images. Given an image $X \in \mathbb{R}^{H \times W \times C}$, models must predict bounding polygons or horizontal segmentation masks $S = \{s_1, s_2, \dots, s_m\}$ mapping to consecutive textual lines. This task is quantified using Intersection over Union (IoU) metrics. It explicitly benchmarks a model's structural resilience against non standardized ruled paper lines, irregular vertical line drift, inter line word bleeding, and heavy erasures or crossings out characteristic of school age writers \cite{Ayadi2025CNN}.

\subsection{End-to-End Paragraph Recognition}
The End-to-End Paragraph Recognition task is a full-page benchmark for educational document understanding. Unlike line- or word-level settings (e.g., KHATT \cite{Mahmoud2014KHATT}), models process entire multi-paragraph pages, requiring reading-order inference, layout understanding, and robustness to noise. Outputs are evaluated as full-page transcripts in markdown or plain text, with CER/WER computed over 100–400 words/page.

%% file: sections/models.tex
To characterize children's Arabic essay recognition, we benchmark 34 models across two groups: (1) off-the-shelf systems evaluated under a unified zero-shot, prompt-conditioned protocol, and (2) our domain-adapted models trained via a two-stage LoRA recipe (\S\ref{ssec:finetuned}).

\subsection{Benchmarked Models}\label{ssec:benchmarked}
All baselines use deterministic decoding with an identical task prompt (\S\ref{sec:gen_benchmark_prompt}) to ensure exact reproducibility.

\paragraph{General-Purpose VLMs.}
We evaluate leading families to study multimodal scaling on authentic handwritten Arabic: \textbf{Qwen2.5-VL} \citep{Qwen2024} and \textbf{Qwen3-VL} \citep{Qwen3VL2025} across 2B--4B, 7B--8B, 32B, and MoE scales (30B-A3B, 235B-A22B); \textbf{Gemma-3} \citep{Gemma2025} at 4B, 12B, scales; and \textbf{Gemma-4}\footnote{\url{https://deepmind.google/technologies/gemma}} at 26B-A4B-it and 31B-it scales. This sweep isolates capacity, architecture, and pretraining language prior effects.
\paragraph{Document-Specialized OCRs.}
We evaluate layout-aware frameworks trained under dense supervision: \textbf{olmOCR} (7B-Preview, 2-7B) \citep{Poznanski2025olmOCR}; \textbf{PaddleOCR-VL} (base, v1.5) \citep{Cui2025PaddleOCRVL}; \textbf{dots.ocr} \citep{Li2025DotsOCR}, which couples layout with sequence structures; and Zhipu AI's \textbf{GLM-OCR} \citep{VTeam2025GLM41V}.

\paragraph{Arabic-Specialized Models.}
We test models optimized for Arabic script properties: \textbf{Baseer} \citep{Hennara2025Baseer} (3B, SLERP-interpolated LoRA); \textbf{Qari-OCR} (v0.3-2B, v0.4-4B) \citep{Wasfy2025QariOCR}, fine-tuned on diacritized text lines; \textbf{AIN-7B} \citep{Heakl2025AIN} (3.6M bilingual samples); QCRI's \textbf{Fanar} (Oryx configuration) \citep{team2025fanar}; and \textbf{Ketaba-OCR-LoRA} \citep{nakbanlp2026ketaba} from NakbaNLP 2026.

\subsection{Domain-Adapted Fine-Tuned Models}\label{ssec:finetuned}
To map performance headroom, we apply a two-stage LoRA recipe to \textbf{Baseer (3B)}, \textbf{Qwen3-VL-4B}, and \textbf{Gemma-4 (27B MoE)}. Projection layers of all Transformer blocks are adapted in \texttt{bfloat16} under a cosine schedule. Dev-set validated checkpoints and hyperparameters are detailed in \S\ref{sec:training_hyperparams}.

\paragraph{Stage~1: Target-Domain Specialization (v1).}
Models are fine-tuned on ArEssays training split with high-capacity LoRA weights and aggressive augmentations, adapting the vision encoder to the irregular strokes, erratic spacing, and varying alignments of child handwriting (Table~\ref{tab:hparams_stage1}).

\paragraph{Stage~2: Curricular Regularization (v2).}
To mitigate catastrophic forgetting and stabilize visual layouts, Stage-1 weights are merged into the base. New lower-rank adapters ($r{=}32$, scaled-down learning rate) are trained on a heterogeneous corpus blending adult script from KHATT \citep{Mahmoud2014KHATT} and printed text from Books-OCR \citep{seemorg_books_ocr}, reducing augmentation ($0.50 \rightarrow 0.20$) to prevent unconstrained text hallucinations (Table~\ref{tab:hparams_stage2}).

\paragraph{Design Rationale.}
The curriculum explicitly decouples \emph{visual specialization} (Stage 1) from \emph{linguistic regularization} (Stage 2). Resolving morphological stroke variance prior to anchoring the distribution with broader canonical text distributions suppresses hallucinations, yielding vastly superior cross-grade generalization over single-stage alternatives.

%% file: sections/results.tex
\definecolor{heatGreen}{HTML}{2E6F40}    %
\definecolor{heatLightG}{HTML}{E2F0D9}   %
\definecolor{heatMidG}{HTML}{C5E1A5}     %
\definecolor{heatLightR}{HTML}{FCE4D6}   %
\definecolor{heatRed}{HTML}{C00000}      %

Empirical evaluations reveal critical architectural disparities, underscoring challenges that developmental Arabic scripts pose to modern VLMs. Table~\ref{tab:benchmark_results_wer_only} summarizes the macro metrics.

\begin{table*}[t]
\centering
\small
\caption{Official Macro Evaluation Leaderboard on the ACHOB Test Split (WER Metrics with Heatmap Gradient).}
\label{tab:benchmark_results_wer_only}
\begin{tblr}{
  colspec = {l c c c},
  row{1} = {font=\bfseries},
  cell{3}{3} = {bg=heatLightR!45}, cell{3}{4} = {bg=heatLightR!35},
  cell{4}{3} = {bg=heatRed!25},   cell{4}{4} = {bg=heatRed!20},
  cell{5}{3} = {bg=heatRed!20},   cell{5}{4} = {bg=heatRed!25},
  cell{6}{3} = {bg=heatRed!30},   cell{6}{4} = {bg=heatRed!35},
  cell{7}{3} = {bg=heatRed!35},   cell{7}{4} = {bg=heatRed!30},
  cell{8}{3} = {bg=heatRed!30},   cell{8}{4} = {bg=heatRed!55},
  cell{10}{3} = {bg=heatLightG!80}, cell{10}{4} = {bg=heatLightG!80},
  cell{11}{3} = {bg=heatLightG!55}, cell{11}{4} = {bg=heatLightG!60},
  cell{12}{3} = {bg=heatLightG!50}, cell{12}{4} = {bg=heatLightG!55},
  cell{13}{3} = {bg=heatLightR!15}, cell{13}{4} = {bg=heatLightR!20},
  cell{14}{3} = {bg=heatLightR!25}, cell{14}{4} = {bg=heatLightR!22},
  cell{15}{3} = {bg=heatLightR!22}, cell{15}{4} = {bg=heatLightR!25},
  cell{16}{3} = {bg=heatLightR!35}, cell{16}{4} = {bg=heatLightR!30},
  cell{17}{3} = {bg=heatLightR!45}, cell{17}{4} = {bg=heatLightR!40},
  cell{18}{3} = {bg=heatRed!15},   cell{18}{4} = {bg=heatRed!15},
  cell{19}{3} = {bg=heatRed!18},   cell{19}{4} = {bg=heatRed!20},
  cell{20}{3} = {bg=heatRed!18},   cell{20}{4} = {bg=heatRed!25},
  cell{21}{3} = {bg=heatRed!25},   cell{21}{4} = {bg=heatRed!30},
  cell{22}{3} = {bg=heatRed!30},   cell{22}{4} = {bg=heatRed!35},
  cell{23}{3} = {bg=heatRed!35},   cell{23}{4} = {bg=heatRed!40},
  cell{24}{3} = {bg=heatRed!35},   cell{24}{4} = {bg=heatRed!38},
  cell{25}{3} = {bg=heatRed!40},   cell{25}{4} = {bg=heatRed!50},
  cell{27}{3} = {bg=heatLightR!55}, cell{27}{4} = {bg=heatLightR!60},
  cell{28}{3} = {bg=heatRed!22},   cell{28}{4} = {bg=heatRed!30},
  cell{29}{3} = {bg=heatRed!20},   cell{29}{4} = {bg=heatRed!26},
  cell{30}{3} = {bg=heatRed!24},   cell{30}{4} = {bg=heatRed!28},
  cell{31}{3} = {bg=heatRed!25},   cell{31}{4} = {bg=heatRed!30},
  cell{32}{3} = {bg=heatRed!30},   cell{32}{4} = {bg=heatRed!45},
  cell{34}{3} = {bg=heatGreen!30},  cell{34}{4} = {bg=heatGreen!40},
  cell{35}{3} = {bg=heatMidG!70},   cell{35}{4} = {bg=heatLightG!40},
  cell{36}{3} = {bg=heatLightG!90}, cell{36}{4} = {bg=heatLightR!15},
  cell{37}{3} = {bg=heatRed!22},   cell{37}{4} = {bg=heatRed!18},
  cell{38}{3} = {bg=heatLightR!50}, cell{38}{4} = {bg=heatRed!26},
  cell{39}{3} = {bg=heatRed!18},   cell{39}{4} = {bg=heatRed!30},
}
\toprule
\textbf{Model Identifier} & \textbf{Category} & \textbf{Exact WER} & \textbf{Corrected WER} \\
\midrule
\textit{Specialized Document OCR Engines} & & & \\
olmOCR-2-7B-1025 & OCR Tool & 74.0\% & 70.4\% \\
PaddleOCR-VL-1.5 & OCR Tool & 101.7\% & 94.2\% \\
olmOCR-7B-0225-preview & OCR Tool & 91.5\% & 102.5\% \\
PaddleOCR-VL & OCR Tool & 115.6\% & 126.0\% \\
dots\_ocr & OCR Tool & 121.3\% & 112.1\% \\
GLM-OCR & OCR Tool & 115.4\% & 160.8\% \\
\midrule
\textit{General Vision-Language Models} & & & \\
Qwen3.6-35B-A3B & Gen VLM & 54.6\% & 52.3\% \\
Qwen3.5-35B-A3B & Gen VLM & 59.7\% & 58.1\% \\
gemma-4-31B-it & Gen VLM & 60.1\% & 59.1\% \\
Qwen3.5-27B & Gen VLM & 64.5\% & 63.1\% \\
Qwen3.6-27B & Gen VLM & 68.7\% & 66.1\% \\
Qwen3.5-9B & Gen VLM & 66.2\% & 68.1\% \\
gemma-4-26B-A4B-it & Gen VLM & 76.0\% & 74.1\% \\
Qwen3.5-4B & Gen VLM & 81.8\% & 80.1\% \\
Qwen3-VL-8B-Instruct & Gen VLM & 92.9\% & 91.5\% \\
Qwen2.5-VL-3B-Instruct & Gen VLM & 97.7\% & 101.2\% \\
gemma-3-12b-it & Gen VLM & 96.7\% & 112.1\% \\
gemma-3-4b-it & Gen VLM & 113.6\% & 122.1\% \\
Qwen3.5-2B & Gen VLM & 125.4\% & 135.1\% \\
Qwen3-VL-4B-Instruct & Gen VLM & 128.1\% & 141.2\% \\
Qwen3-VL-2B-Instruct & Gen VLM & 129.3\% & 138.1\% \\
Qwen2.5-VL-7B-Instruct & Gen VLM & 135.2\% & 154.1\% \\
\midrule
\textit{Arabic Regional Specialists} & & & \\
Ketaba-OCR-LoRA & Arabic Spec & 88.1\% & 89.6\% \\
FanarOryx & Arabic Spec & 105.4\% & 123.1\% \\
Baseer\_\_Nakba & Arabic Spec & 102.5\% & 115.4\% \\
Qari-OCR-0.4.0-VL-4B-Instruct & Arabic Spec & 109.8\% & 118.5\% \\
Qari-OCR-v0.3-VL-2B-Instruct & Arabic Spec & 110.4\% & 121.4\% \\
AIN & Arabic Spec & 119.9\% & 146.9\% \\
\midrule
\textit{Domain-Adapted Fine-Tuned (Ours)} & & & \\
Qwen3.5-4B (FT Stage 1 - v1) & Fine-Tuned & \textbf{46.9\%} & 41.9\% \\
Gemma4-26B-A4B (FT Stage 1 - v1) & Fine-Tuned & 49.2\% & \textbf{64.3\%} \\
Baseer-Nakba (FT Stage 1 - v1) & Fine-Tuned & 51.0\% & 69.1\% \\
Qwen3.5-4B (FT Stage 2 - v2) & Fine-Tuned & 107.2\% & 99.0\% \\
Gemma4-26B-A4B (FT Stage 2 - v2) & Fine-Tuned & 86.8\% & 114.5\% \\
Baseer-Nakba (FT Stage 2 - v2) & Fine-Tuned & 99.4\% & 122.8\% \\
\bottomrule
\end{tblr}
\end{table*}

\subsection{Macro Leaderboard Performance}
\begin{figure}[h]
\centering
\includegraphics[width=0.80\linewidth]{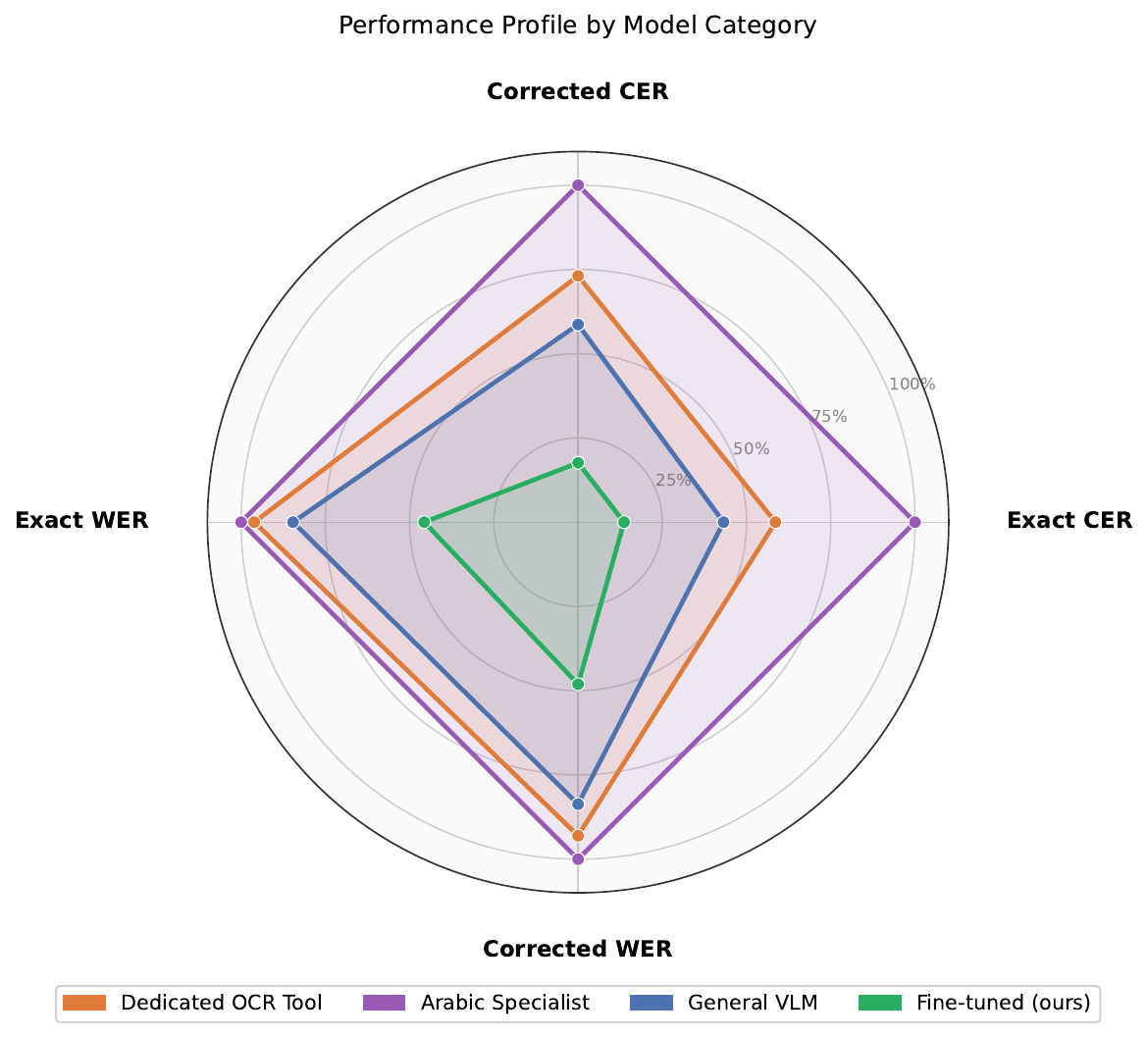}
\caption{Cross category performance profiles across error metrics. Lower normalized area indicates optimal baseline cross task stability.}
\label{fig:performance_radar}
\end{figure}

Parameter scaling reduces zero-shot errors, with dense Qwen variants yielding major gains. Conversely, document-centric layouts like \texttt{dots\_ocr} or \texttt{GLM-OCR} fail on distorted text lines, proving that handwritten morphic variance requires explicit domain-adaptive mechanisms over standard layout parsing.

\subsection{The \texorpdfstring{$\Delta$}{Delta}CER Performance Gap}
The variance between verbatim transcriptions (\texttt{exact\_cer}) and language-smoothed outputs (\texttt{corrected\_cer}) flags a strong textual prior within open VLMs:
$$\Delta\text{CER} = \text{exact\_cer\_mean} - \text{corrected\_cer\_mean}$$
Unadapted architectures like Qwen3.6-35B-A3B exhibit a negative gap ($\Delta\text{CER} \approx -0.07$). When facing noisy strokes or missing diacritics (\textit{I'jam}), these systems utilize internal auto-regressive distributions to output standard literary phrasings, sacrificing literal token accuracy. Document OCR tools track near zero, indicating rigid extraction profiles.

\subsection{Impact of Human Annotated Image Quality Factors}
\begin{figure}[h]
\centering
\includegraphics[width=0.95\linewidth]{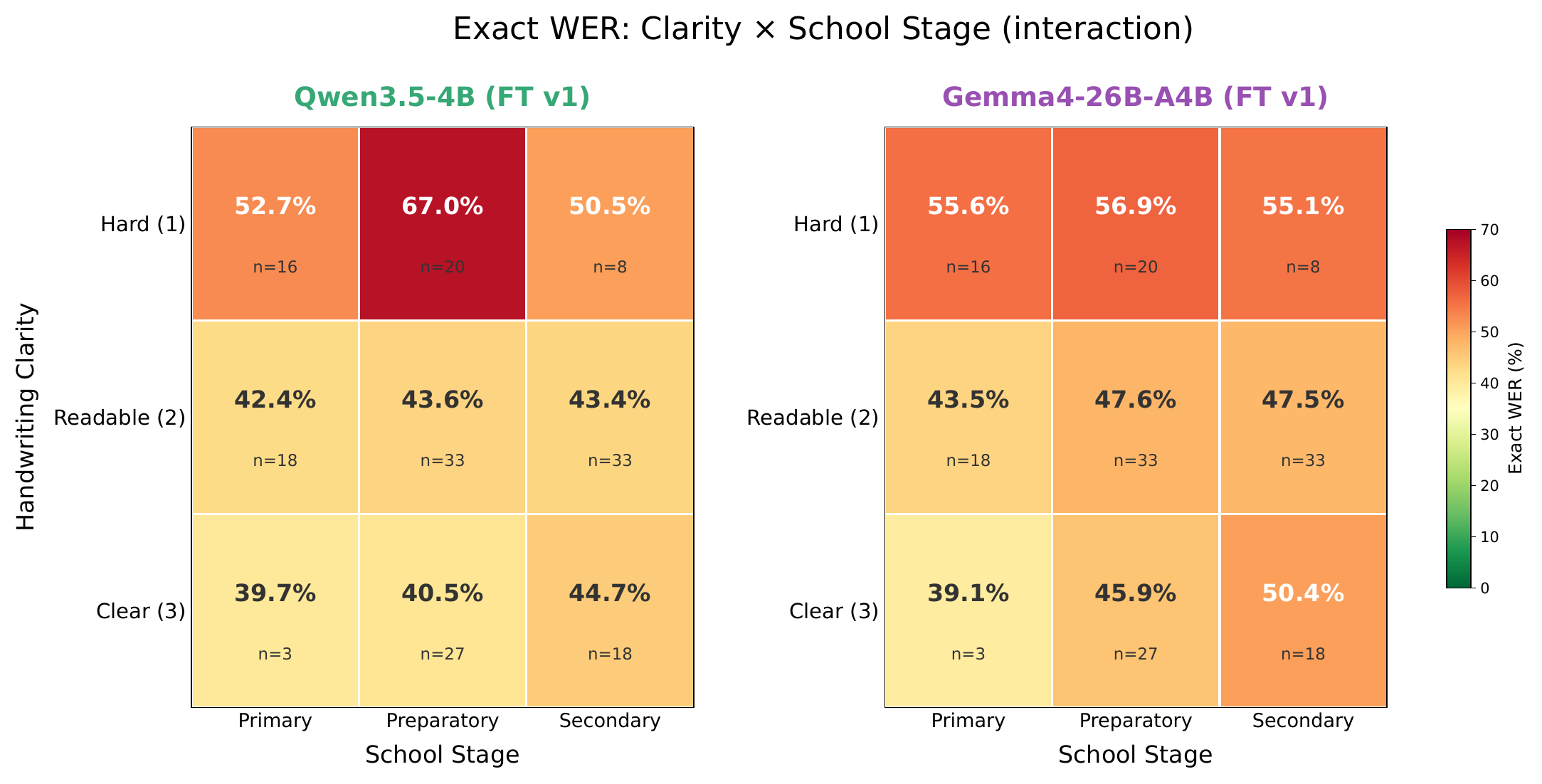}
\caption{Impact of handwriting clarity and school stage interactions on WER. Low handwriting clarity acts as the primary bottleneck across cohorts.}
\label{fig:clarity_heatmap}
\end{figure}

Stratifying metrics reveals that handwriting clarity governs sequence accuracy: illegible scripts induce severe base model hallucinations, whereas lighting noise causes minimal errors. This confirms vision encoders tolerate imaging artifacts but struggle with structural character variations.

\subsection{Efficacy of Domain Adapted Fine Tuned Specialists}
Targeted fine-tuning yields immense verbatim calibration. Stage 1 (v1) specialists trained on child data lead the benchmark, where \texttt{qwen35\_4b\_v1} achieves top macro standing (0.1631 Exact CER), followed by \texttt{gemma4\_26b\_v1}.

Multi-stage regularizers in Stage 2 (v2), combining KHATT adult samples and printed Books-OCR text, introduce a critical regularization tradeoff. While v2 regularizers support standard lines, they degrade specialized resolution on native child writing, increasing \texttt{qwen35\_4b\_v2} errors. This highlights the need for precise curriculum balancing across writer age segments.

\subsection{Developmental Gradient Analysis}
\begin{figure}[h]
\centering
\includegraphics[width=0.95\linewidth]{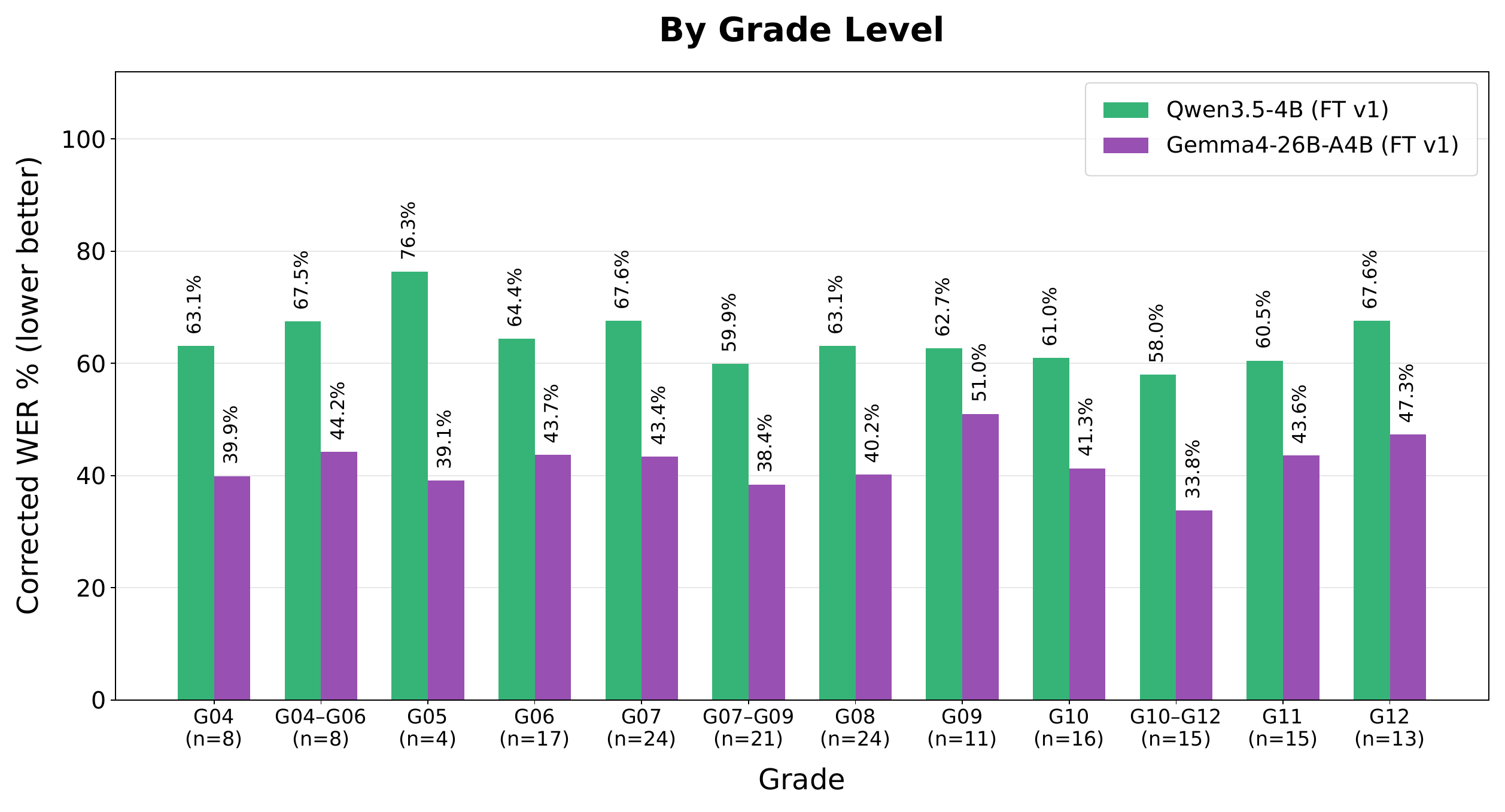}
\caption{Corrected Word Error Rate (WER) across distinct educational tiers on top 2 finetuned models.}
\label{fig:ft_corrected_wer}
\end{figure}
Metrics uncover a clear error gradient tied to motor-spatial development. Primary education scripts (Grades 4 through 6) show peak error density due to unstable baselines and irregular character junctions. As text layout and stroke proportions stabilize in Preparatory and Secondary cohorts, word and character error tracking exhibits a steady linear reduction.